\documentclass[conference]{IEEEtran}

\usepackage{cite}
\usepackage{amsmath,amssymb,amsfonts}
\usepackage{algorithmic}
\usepackage{graphicx}
\usepackage{textcomp}
\usepackage{xcolor}
\usepackage{glossaries}
\usepackage{subcaption}
\usepackage{hyperref}
\usepackage{colortbl}  

\usepackage{booktabs,tabularx,makecell,array}
\newcolumntype{Y}{>{\raggedright\arraybackslash}X}

\def\BibTeX{{\rm B\kern-.05em{\sc i\kern-.025em b}\kern-.08em
    T\kern-.1667em\lower.7ex\hbox{E}\kern-.125emX}}
\newacronym{dfa}{DFA}{Direct Feedback Alignment}
\newacronym{ai}{AI}{Artificial Intelligence}
\newacronym{asic}{ASIC}{Application-Specific Integrated Circuit}
\newacronym{bptt}{BPTT}{Back-Propagation Through Time}
\newacronym{bram}{BRAM}{Block RAM}
\newacronym{cl}{CL}{Continual Learning}
\newacronym{cpu}{CPU}{Central Processing Unit}
\newacronym{fpga}{FPGA}{Field Programmable Gate Array}
\newacronym{hdl}{HDL}{Hardware Description Language}
\newacronym{hw}{HW}{Hardware}
\newacronym{lut}{LUT}{Look Up Table}
\newacronym{mac}{MAC}{Multiply and Accumulate}
\newacronym{ml}{ML}{Machine Learning}
\newacronym{mlp}{MLP}{Multi-Layer Perceptron}
\newacronym{nn}{NN}{Neural Network}
\newacronym{snn}{SNN}{Spiking Neural Network}
\newacronym{stsf}{STSF}{Spiking Time Sparse Feedback}
\newacronym{mse}{MSE}{Mean Squared Error}

\newacronym{adse}{ADSE}{Automatic Design Space Exploration}
\newacronym{aer}{AER}{Address Event Representation}
\newacronym{ann}{ANN}{Artificial Neural Network}
\newacronym{api}{API}{Application Programming Interface}
\newacronym{ax}{AX}{Adaptive eXperimentation}
\newacronym{cnn}{CNN}{Convolutional Neural Network}
\newacronym{cots}{COTS}{Commercial Off The Shelf}
\newacronym{csnn}{CSNN}{Convolutional Spiking Neural Network}
\newacronym{cu}{CU}{Control Unit}
\newacronym{dnn}{DNN}{Deep Neural Network}
\newacronym{dp}{DP}{Data Path}
\newacronym{dram}{DRAM}{Dynamic Random Access Memory}
\newacronym{dse}{DSE}{Design Space Exploration}
\newacronym{dvs}{DVS}{Dynamic Vision Sensor}
\newacronym{eda}{EDA}{Electronic Design Automation}
\newacronym{elut}{ELUT}{Equivalent Look Up Table}
\newacronym{eth}{ETH}{Eidgenössische Technische Hochschule}
\newacronym{fc}{FC}{Fully-Connected}
\newacronym{fcr}{FC-R}{Fully-Connected Recurrent}
\newacronym{ff}{FF}{Flip Flop}
\newacronym{fffc}{FF-FC}{Feed-Forward Fully-Connected}
\newacronym{FFNN}{FFNN}{Feed-Forward Neural Network}
\newacronym{fi}{FI}{Fault Injection}
\newacronym{fim}{FIM}{Fault Injection Manager}
\newacronym{fl}{FL}{Fault List}
\newacronym{flg}{FLG}{Fault List Generator}
\newacronym{fsm}{FSM}{Finite State Machine}
\newacronym{gpgpu}{GPGPU}{General Purpose Graphic Processing Unit}
\newacronym{gpu}{GPU}{Graphic Processing Unit}
\newacronym{if}{IF}{Integrate and Fire}
\newacronym{ini}{INI}{Institute of Neuro-Informatics}
\newacronym{iot}{IoT}{Internet of Things}
\newacronym{lif}{LIF}{Leaky Integrate and Fire}
\newacronym{lr}{LR}{Latent Replay}
\newacronym{lstm}{LSTM}{Long Short Term Memory}
\newacronym{nas}{NAS}{Network Architecture Search}
\newacronym{ne}{NE}{Network Evaluator}
\newacronym{ng}{NG}{Network Generator}
\newacronym{nir}{NIR}{Neuromorphic Intermediate Representation}
\newacronym{nlp}{NLP}{Natural Language Processing}
\newacronym{ostl}{OSTL}{Online Spatio-Temporal Learning}
\newacronym{ostp}{OSTP}{Online Spatio-Temporal Learning with Target Projection}
\newacronym{pu}{PU}{Processing Unit}
\newacronym{pulp}{PULP}{Parallel processing Ultra-Low Power}
\newacronym{qat}{QAT}{Quantization Aware Training}
\newacronym{ram}{RAM}{Random Access Memory}
\newacronym{rcr}{RC-R}{Randomly-Connected Recurrent}
\newacronym{rl}{RL}{Reinforcement Learning}
\newacronym{rnn}{RNN}{Recurrent Neural Network}
\newacronym{rom}{ROM}{Read Only Memory}
\newacronym{rsnn}{RSNN}{Recurrent Spiking Neural Network}
\newacronym{rtl}{RTL}{Register Transfer Level}
\newacronym{rtrl}{RTRL}{Real-Time Recurrent Learning}
\newacronym{sbs}{SbS}{Spike-by-Spike}
\newacronym{scnn}{SCNN}{Spiking Convolutional Neural Networks}
\newacronym{sfi}{SFI}{Statistical Fault Injection}
\newacronym{sg}{SG}{Surrogate Gradient}
\newacronym{shd}{SHD}{Spiking Heidelberg Dataset}
\newacronym{simd}{SIMD}{Single Instruction Multiple Data}
\newacronym{soc}{SoC}{System on Chip}
\newacronym{sota}{SOTA}{State Of The Art}
\newacronym{sram}{SRAM}{Static Random Access Memory}
\newacronym{srm}{SRM}{Spike Response Model}
\newacronym{stdp}{STDP}{Spike-Timing-Dependent Plasticity}
\newacronym{tpu}{TPU}{Tensor Processing Unit}
\newacronym{ucb}{UCB}{Upper Confidence Bound}
\newacronym{vhdl}{VHDL}{VHSIC Hardware Description Language}
\newacronym{wta}{WTA}{Winner Takes All}
\newacronym{ANN}{ANN}{Artificial Neural Network}
\newacronym{PNN}{PNN}{Photonic Neural Network}
\newacronym{ADC}{ADC}{Analog-to-Digital Converter}
\newacronym{DAC}{DAC}{Digital-to-Analog Converter}
\newacronym{MZI}{MZI}{Mach-Zehnder Interferometer}
\newacronym{MVM}{MVM}{Matrix-Vector Multiplication}
\newacronym{CMOS}{CMOS}{Complementary Metal-Oxide-Semiconductor}
\newacronym{ASIC}{ASIC}{Application-Specific Integrated Circuit}
\newacronym{SVD}{SVD}{Singular Value Decomposition}
\newacronym{epsp}{EPSP}{Excitatory Post-Synaptic Potential}
\newacronym{ltp}{LTP}{Long-Term Potentiation}
\newacronym{ltd}{LTD}{Long-Term Depression}
\newacronym{mzi}{MZI}{Mach-Zehnder Interferometers}
\newacronym{ptp}{PTP}{Post-Tetanic Potentiation}

\newcommand{\tool}{\textsc{Spiker-LL}}

\begin{document}

\title{\tool: An Energy-Efficient FPGA Accelerator Enabling Adaptive Local Learning in Spiking Neural Networks\\
\thanks{Identify applicable funding agency here. If none, delete this.}
}

\author{\IEEEauthorblockN{Alessio Caviglia, Filippo Marostica, Alessandro Savino, Stefano Di Carlo}
\IEEEauthorblockA{Control and Computer Engineering Department, Politecnico di Torino, Turin, Italy\\
Email: \{alessio.caviglia, filippo.marostica, alessandro.savino, stefano.dicarlo\}@polito.it}
}


\maketitle

\begin{abstract}

  Deploying adaptive intelligence at the edge remains challenging due to the high computational and energy cost of training neural models. \glspl{snn} offer a promising alternative, but enabling on-device learning requires hardware–algorithm co-design. This paper presents \tool{}, an FPGA-based \gls{snn} accelerator that extends the open-source Spiker+ inference architecture with efficient support for the STSF local learning rule. Through targeted microarchitectural extensions, \tool{} performs inference and online learning with minimal overhead. Across MNIST, F-MNIST, and DIGITS, it achieves up to 93\% accuracy, sub-millisecond latency, and $<$0.1 mJ per inference, while remaining DSP-free and highly scalable for edge-FPGA deployments.
\end{abstract}

\begin{IEEEkeywords}
neuromorphic computing, spiking neural networks (SNNs), hardware accelerator, on-chip learning, local learning rules
\end{IEEEkeywords}

\section{Introduction}
\glsresetall

The rapid progress of \gls{ml} has made adaptive intelligent systems increasingly feasible, but training remains computationally and energy intensive~\cite{schwartz2020green}, a critical limitation for edge devices where every milliwatt matters. Within this broader trend, progress in \gls{snn} training algorithms and hardware accelerators design is making neuromorphic edge intelligence increasingly practical. \glspl{snn}~\cite{maass1997networks} exploit event-driven, temporally sparse computation and can solve learning tasks under tight resource constraints~\cite{11130320}, yet a major challenge remains: enabling learning directly at the edge.
This requires tight co-design between training algorithms and hardware. While full \gls{bptt} \cite{58337}, supported by frameworks such as snnTorch~\cite{eshraghian2023training}, remains a gold standard for \glspl{snn}, it is typically infeasible on resource-constrained devices~\cite{58337}. This has motivated local learning rules, which compute updates where and when data is generated, reducing computation and memory demands. At the same time, general-purpose CPUs/GPUs are often too slow or power-hungry for event-driven spiking workloads, which has driven the development of dedicated \gls{snn} accelerators across \gls{asic} and \gls{fpga} platforms. Here, we focus on \glspl{fpga} because they combine low deployment cost with reconfigurability, a practical advantage in a rapidly evolving design space.

This paper presents \tool{}, an \gls{fpga}-based \gls{snn} accelerator that extends the open-source Spiker architecture~\cite{carpegnaSpikerFrameworkGeneration2024,9911998, sfatti2025} with efficient, fully on-device learning. While Spiker previously supported only inference, \tool{} introduces hardware mechanisms for on-device training, broadening the platform toward adaptive edge intelligence.
Our contributions are threefold. (1) \textbf{Algorithm--hardware integration}: \tool{} implements the \gls{stsf} local learning rule~\cite{heSTSFSpikingTime2025}, enabling supervised learning without the cost of full \gls{bptt}. (2) \textbf{Microarchitectural extensions for on-device learning}: training support is added through minimal, targeted modifications at synaptic-state access points, while reusing existing datapaths, memory banks, and control logic to preserve Spiker's timing closure and efficiency. (3) \textbf{A unified inference--training accelerator}: \tool{} supports real-time inference and on-device learning in the same hardware pipeline under tight energy and area constraints, enabling deployment on low-cost edge \glspl{fpga}.

Across MNIST, F-MNIST, and DIGITS, \tool{} achieves up to 92--93\% accuracy while maintaining sub-millisecond latency and energy per inference below 0.1\,mJ. Despite adding training capabilities, the design remains DSP-free and scales from ultra-compact configurations with less than 5k \glspl{lut} to larger networks without compromising real-time operation. Compared to prior work, \tool{} offers competitive accuracy, low energy per inference, and a flexible design well suited to resource-constrained edge FPGAs.

\section{Background}
\label{sec:background}

This section reviews the \gls{snn} computation model, the hardware limitations of existing training algorithms, and recent advances in local on-device learning.

\subsection{Spiking Neural Networks}

\glspl{snn} compute through discrete, event-driven activations that map naturally to hardware-efficient implementations: neurons emit binary spikes $s(t)\in\{0,1\}$ when their membrane potential exceeds a threshold, enabling sparse communication instead of dense vector operations. Inputs may come from neuromorphic sensors or be encoded into spike trains through rate, temporal, or latency coding~\cite{maass1997networks}. Most digital accelerators adopt the discrete-time \gls{lif} model, consisting of synaptic accumulation followed by a recurrent membrane update:

\begin{equation}
\small
\begin{aligned}
I_{\mathrm{syn}}[n] &= \sum_j W_j\, s_{\mathrm{in},j}[n], \\
V_m[n] &= \beta V_m[n-1] + I_{\mathrm{syn}}[n]
         - V_{\mathrm{th}} s_{\mathrm{out}}[n-1], \\
s_{\mathrm{out}}[n] &=
\begin{cases}
1, & V_m[n] > V_{\mathrm{th}},\\
0, & \text{otherwise}
\end{cases}
\end{aligned}
\label{eq:lif}
\end{equation}

where $W_j$ are synaptic weights, $\beta=e^{-\Delta t/\tau_m}$ is the membrane decay factor, $\tau_m$ the membrane time constant, $V_m[n]$ the membrane potential, and $I_{\mathrm{syn}}[n]$ the synaptic input. A spike is emitted when $V_m$ crosses $V_{\mathrm{th}}$, after which the potential is reset. These equations define the core datapath of digital \gls{snn} accelerators: weighted accumulation, recurrent state update, and threshold--reset logic, typically pipelined to meet timing constraints.

\subsection{Training Spiking Neural Networks}

A major obstacle to efficient \gls{snn} training is the non-differentiable spike function. Software frameworks such as snnTorch~\cite{eshraghian2023training} address this using \gls{bptt}~\cite{58337} with surrogate gradients~\cite{8891809}. However, \gls{bptt} requires storing and backpropagating through $N$ neuron states over $T$ timesteps, incurring $\mathcal{O}(NT)$ memory and $\mathcal{O}(ET)$ compute cost, where $E$ is the number of synapses. This temporal unrolling creates long-range dependencies and irregular dataflows that conflict with the streaming, feed-forward execution model of edge accelerators~\cite{eshraghian2023training}. As a result, \gls{bptt} breaks timestep independence, requires non-local error communication, demands large state buffers, and introduces feedback paths misaligned with accelerator pipelines, making it unsuitable for low-power edge devices.

To enable hardware-feasible training, recent work has shifted to \emph{local}, \emph{on-device} learning rules that compute updates using only information available at the current timestep. These rules avoid temporal unrolling and reduce memory to synapse-local state already present during inference. Representative examples include e-Prop~\cite{bellec2020solution}, which maintains on-device eligibility traces and neuron-local error signals; DECOLLE~\cite{10.3389/fnins.2020.00424}, which uses layer-wise local random readouts to generate surrogate losses; and \gls{stsf}~\cite{heSTSFSpikingTime2025}, which uses \gls{dfa} to provide fixed random top-down error signals and applies spike-triggered local plasticity. These methods share an architectural advantage: synaptic updates depend only on local variables (e.g., spikes, membrane potential, eligibility traces) and, at most, a broadcast error signal, eliminating global backpropagation.
We adopt \gls{stsf} as the learning rule for our accelerator because it offers an effective balance between learning performance and hardware efficiency. Unlike trace-based methods such as e-Prop or DECOLLE, which require per-synapse/neuron eligibility traces and additional temporal state, \gls{stsf} uses only spike timing and local pre/post activity, enabling event-driven weight updates with negligible state overhead. This removes the need for large trace memories and decay circuitry, significantly reducing hardware cost and power, especially at scale. Although \gls{stsf} provides more limited temporal credit assignment and may underperform on tasks with long-range dependencies, it is well suited for feedforward, low-latency spiking workloads. In these scenarios, it offers a practical compromise that keeps on-chip learning lightweight while preserving the advantages of neuromorphic computation.

\subsection{Hardware Accelerators for Training SNNs}

The emergence of simple local learning rules is essential for enabling spiking intelligence at the edge, but it is not sufficient: even local updates require per-synapse operations, error modulation, and state maintenance that remain costly on CPUs and microcontrollers. Real-time on-device learning therefore demands dedicated hardware with predictable latency, low energy, and tight memory use. Large neuromorphic systems such as Intel's Loihi support reward-modulated \gls{stdp}~\cite{8259423}, but are proprietary and unsuitable for low-cost, customizable edge deployments. Other hardware efforts include \gls{asic} designs such as ReckOn, which implements e-Prop~\cite{frenkelReckOn28nmSubmm22022}, and \gls{fpga} designs that explore different trade-offs in energy, accuracy, and learning mechanisms~\cite{11208669,10841405,9822407,10492616,thaon.FPGABasedCoProcessorSpiking2022,leeSpikeTrainLevelDirect2020,siddiqueLowCostNeuromorphic2023}.
Despite this progress, open-hardware support for supervised local rules remains very limited. By implementing \gls{stsf} on the open-source Spiker \gls{fpga} architecture~\cite{carpegnaSpikerFrameworkGeneration2024}, our work helps fill this gap and provides an accessible, configurable platform for studying supervised on-device learning directly in hardware—an important step toward deployable, adaptive edge intelligence.

\section{Methods}

\begin{figure*}[htbp]
  \centering
  \includegraphics[width=0.95\textwidth, trim=0cm 0.97cm 0cm 0cm,
                   clip]{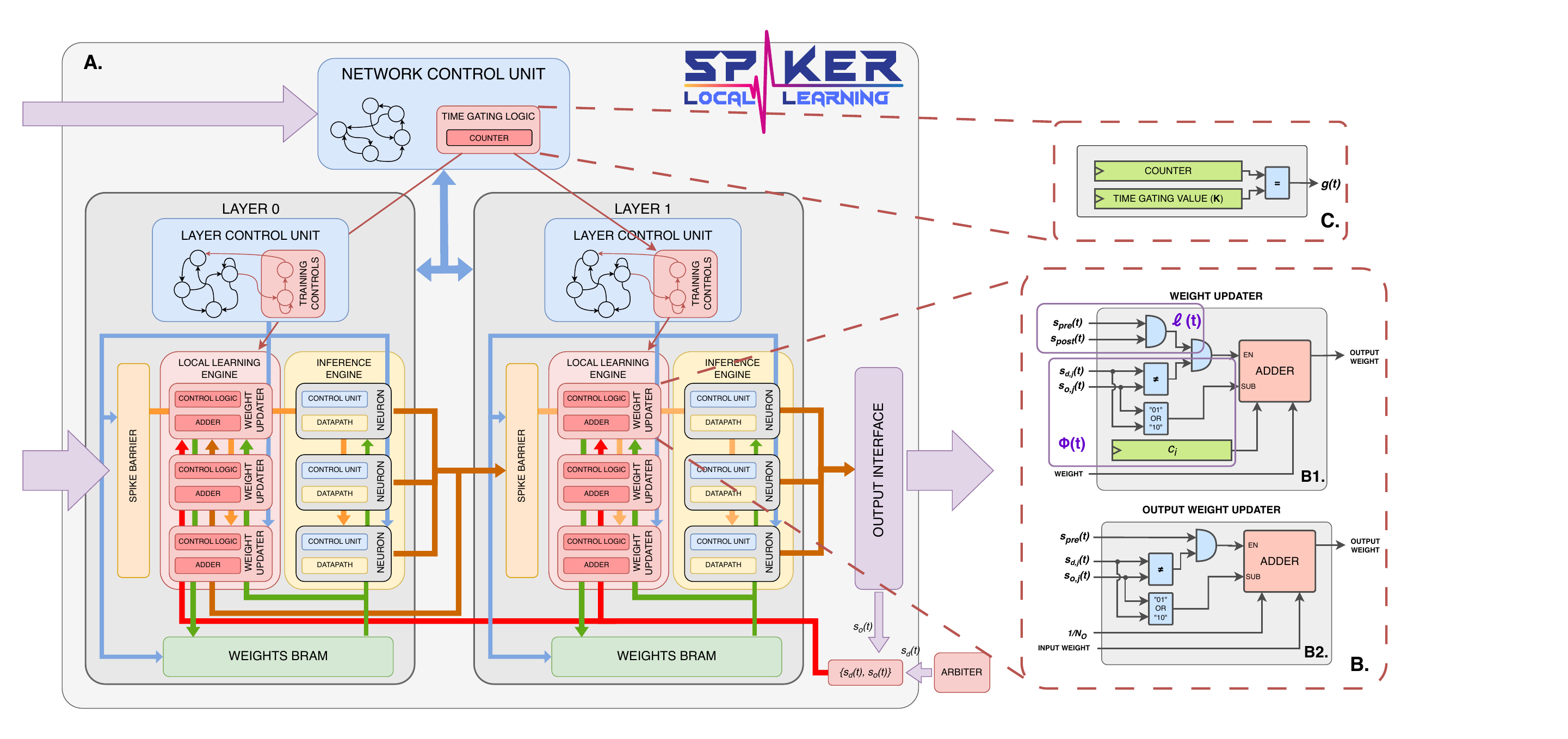}
  \caption{Global view of the \tool{} architecture.
\tool{} extends the Spiker+ inference-oriented architecture with additional local and global structures that implement and coordinate the on-device learning process.}
  \label{fig:overview}
\end{figure*}

Figure~\ref{fig:overview}-A shows a high-level view of \tool{}. The design extends the fully parameterizable, highly optimized Spiker+ inference accelerator with dedicated hardware for supervised on-device learning. The baseline architecture consists of fully connected layers of multiplier-free \gls{lif} neurons with configurable reset mechanisms~\cite{carpegnaSpikerFrameworkGeneration2024}. Each layer is managed by a local controller handling synaptic accumulation, membrane updates, and spike generation, while a global controller coordinates inter-layer sequencing.
Weights are stored in per-layer \glspl{bram}, enabling low-energy, locality-aware access. Inputs arrive as digital spike streams injected at the first-layer barrier, with spike encoding handled off-chip. Neurons compute in parallel, while input channels are streamed sequentially to match \gls{fpga} routing and \gls{bram} port constraints.
To support learning, \tool{} adds lightweight local learning modules (highlighted in red in Figure~\ref{fig:overview}) that implement the synaptic update rules of \gls{stsf}~\cite{heSTSFSpikingTime2025}. These modules integrate with the existing datapath, control logic, and memory system while preserving the performance and resource efficiency of Spiker+.

\subsection{Learning architecture}

\subsubsection{Local learning module}

\tool{} equips each \gls{lif} neuron with an ultra-lightweight, tightly coupled weight updater (Figure~\ref{fig:overview}--B), implementing the \gls{stsf} three-factor rule directly in the accelerator datapath. This modular solution can be deployed across different network configurations. For each synapse, the weight update combines three quantities available in hardware: (i) the pre-synaptic spike $s_{\mathrm{pre}}(t)$, (ii) the post-synaptic spike $s_{\mathrm{post}}(t)$, and (iii) a global modulatory signal $\Phi(t)$ distributed to all neurons (Figure~\ref{fig:overview}-A). The rule is
\begin{equation}
\small
\Delta w = \sum_{t} \Phi(t)\,\ell(t),
\label{eq:three_factor}
\end{equation}
where $\ell(t)$ is the local plasticity term computed inside the weight updater:
\begin{equation}
\small
\ell(t) \equiv e\big(s_{\mathrm{pre}}(t),\,s_{\mathrm{post}}(t)\big).
\end{equation}
The local contribution $\ell(t)$ is derived from \gls{stdp}. In its general form, \gls{stdp} requires tracking all pre/post spike-time differences:
\begin{equation}
\small
\Delta w =
\sum_{t_\text{pre}}
\sum_{t_\text{post}}
F\left(t_\text{post}-t_\text{pre}\right),
\label{eq:stdp_general}
\end{equation}
with kernel
\begin{equation}
\small
F(\Delta t) =
\begin{cases}
A_{+}\,e^{\big(-\Delta t / \tau_{+}\big)}, & \Delta t > 0,\\[2pt]
- A_{-}\,e^{\big(\Delta t / \tau_{-}\big)}, & \Delta t < 0,
\end{cases}
\label{eq:stdp_kernel}
\end{equation}
where $A_{+}, A_{-} > 0$ and $\tau_{+}, \tau_{-}$ are time constants. This requires spike traces and time-difference computations, which are expensive in streaming \gls{fpga} datapaths.
To maintain strict locality and avoid trace memories, \tool{} implements a simplified variant (vanilla \gls{stdp}) \cite{heSTSFSpikingTime2025} in which updates depend only on spike co-occurrence within the same discrete timestep:
\begin{equation}
\small
\ell(t) \equiv e\big(s_{\text{pre}}(t),\,s_{\text{post}}(t)\big)
= s_{\text{pre}}(t)\cdot s_{\text{post}}(t).
\label{eq:vanilla_stdp}
\end{equation}
Because spikes are binary, this term is non-zero only when both spikes occur in the same cycle. In Spiker+, pre- and post-synaptic spikes are simultaneously available at the neuron interface, so $\ell(t)$ is implemented as a simple \texttt{AND} gate enabling the synapse weight accumulator (Figure~\ref{fig:overview}--B1), with no temporal buffers or trace registers.

The modulatory signal $\Phi(t)$ is obtained from a global error combined with \gls{dfa}, which projects the output error through fixed random feedback matrices to generate per-layer learning signals without backward passes. \tool{} uses the derivative of the mean squared error with respect to the output spikes as the global error signal. For an output layer with $N_o$ neurons, the instantaneous loss is
\begin{equation}
\small
L(t) = \frac{1}{N_o}
\sum_{j=1}^{N_o}
\big(s_{d,j}(t) - s_{o,j}(t)\big)^2,
\label{eq:mse_loss}
\end{equation}
where $s_{o,j}(t)$ and $s_{d,j}(t)$ are the actual and desired spikes of output neuron $j$. The corresponding output error term is
\begin{equation}
\small
\delta_j(t) = \frac{\partial L(t)}{\partial s_{o,j}(t)}
= \frac{2}{N_o}\big(s_{o,j}(t) - s_{d,j}(t)\big),
\label{eq:mse_grad}
\end{equation}
which is ternary because $s_{o,j}(t), s_{d,j}(t)\in\{0,1\}$. This is favorable for hardware, since only sign/zero information is needed and constant factors can be absorbed into precomputed gains.
Error propagation to the hidden layer uses \gls{dfa} with a sparse fixed random feedback matrix $\mathbf{B}\in\mathbb{R}^{N_h\times N_o}$, where $N_h$ is the number of hidden neurons. Collecting output errors in $\boldsymbol{\delta}(t)=[\delta_1(t),\dots,\delta_{N_o}(t)]^\top$, the modulatory signal for hidden neuron $i$ is
\begin{equation}
\small
\Phi_i(t) = \sum_{j=1}^{N_o} B_{ij}\,\delta_j(t).
\label{eq:dfa_phi}
\end{equation}
Following STSF, $\mathbf{B}$ is made extremely sparse while avoiding all-zero rows/columns. In our implementation, each row $i$ contains exactly one non-zero entry at column $k(i)$, so
\begin{equation}
\small
\Phi_i(t) = B_{i k(i)}\,\delta_{k(i)}(t),
\label{eq:dfa_phi_simplified}
\end{equation}
and each hidden neuron receives feedback from a single output neuron. Consequently, $\Phi_i(t)$ is ternary:
\begin{equation}
\small
\Phi_i(t) \in \{0,\,+c_i,\,-c_i\},
\end{equation}
where $c_i$ is a precomputed quantized constant for hidden neuron $i$ (including feedback weight and global scaling/learning-rate factors).
Hardware realization exploits this fixed magnitude. For each hidden neuron, \tool{} stores the sign and magnitude of $c_i$ in \glspl{lut} and uses a small control block to decode the error of output neuron $k(i)$, driving the \texttt{EN} and \texttt{SUB} inputs of the local adder to select among $-c_i$, $0$, and $+c_i$. This is combined with the vanilla-STDP term by AND-ing the enable conditions, so a weight update is issued only when both the global error and local spike-coincidence conditions are satisfied (Figure~\ref{fig:overview}--B1).

In the final \tool{} architecture, each layer instantiates a dedicated local learning module with as many weight updaters as neurons in the layer, mirroring the original Spiker+ datapath and enabling fully parallel updates across the neuron population (Figure~\ref{fig:overview}--A).
Output-layer synapses differ because they connect directly to output neurons. Their updaters are therefore simpler (Figure~\ref{fig:overview}--B2): no \gls{dfa} propagation or hidden-layer local STDP term is required. For the synapse from hidden neuron $i$ to output neuron $j$, we use the discrete update rule
\begin{equation}
\small
\Delta w_{ij}
= \eta_{\text{out}} \sum_{t} \delta_j(t)\,s^{(h)}_i(t),
\label{eq:output_update}
\end{equation}
where $s^{(h)}_i(t)$ is the spike of hidden neuron $i$ and $\delta_j(t)$ is the ternary output error. Since $s^{(h)}_i(t)\in\{0,1\}$, updates occur only when neuron $i$ fires, while $\delta_j(t)$ selects potentiation or depression.
As in the hidden layer, the output-layer update is ternary. Here the update magnitude is identical for all synapses, so \tool{} precomputes and quantizes the shared constant inside the output-layer trainer module and uses control logic analogous to the hidden-layer case to select the signed update value (Figure~\ref{fig:overview}--B2).

\subsubsection{Arbiter}
\label{sec:arbiter}

The only non-local information needed by the weight updaters is a per-output-neuron error bit. In our supervised experiments this is derived from the label, but \tool{} does not require supervised targets at runtime. The same one-bit feedback can be generated by an external arbiter, including task-level heuristics, self-supervised signals, or simple reward indicators. This keeps the interface general and compatible with closed-loop online learning using real sensors even when ground-truth labels are unavailable.

\subsubsection{Control units}

Spiker's control units were minimally extended to support training. As shown in Fig.~\ref{fig:overview}, each per-layer \gls{fsm} adds training states that mirror inference and reuse the same control signals. To mitigate overfitting and reduce training time, we introduce temporal gating in the \gls{stsf} rule: a small counter with a runtime-configurable value enables weight updates only every $K$ timesteps (Figure~\ref{fig:overview}-C). With gating signal $g(t)$, the effective update becomes
\begin{equation}
\small
\Delta w = \sum_{t} g(t)\,\Phi(t)\,e\big(\text{pre}(t),\,\text{post}(t)\big),
\label{eq:gated_update}
\end{equation}
with $g(t)=1$ only when learning is enabled (e.g., $g(t)=1$ if $t \bmod K = 0$, and $0$ otherwise). Updates are applied immediately at each qualifying timestep in the same fixed-point format used for inference, with no separate accumulation buffer.

\begin{figure}[htbp]
  \centering
  \includegraphics[scale=0.32]{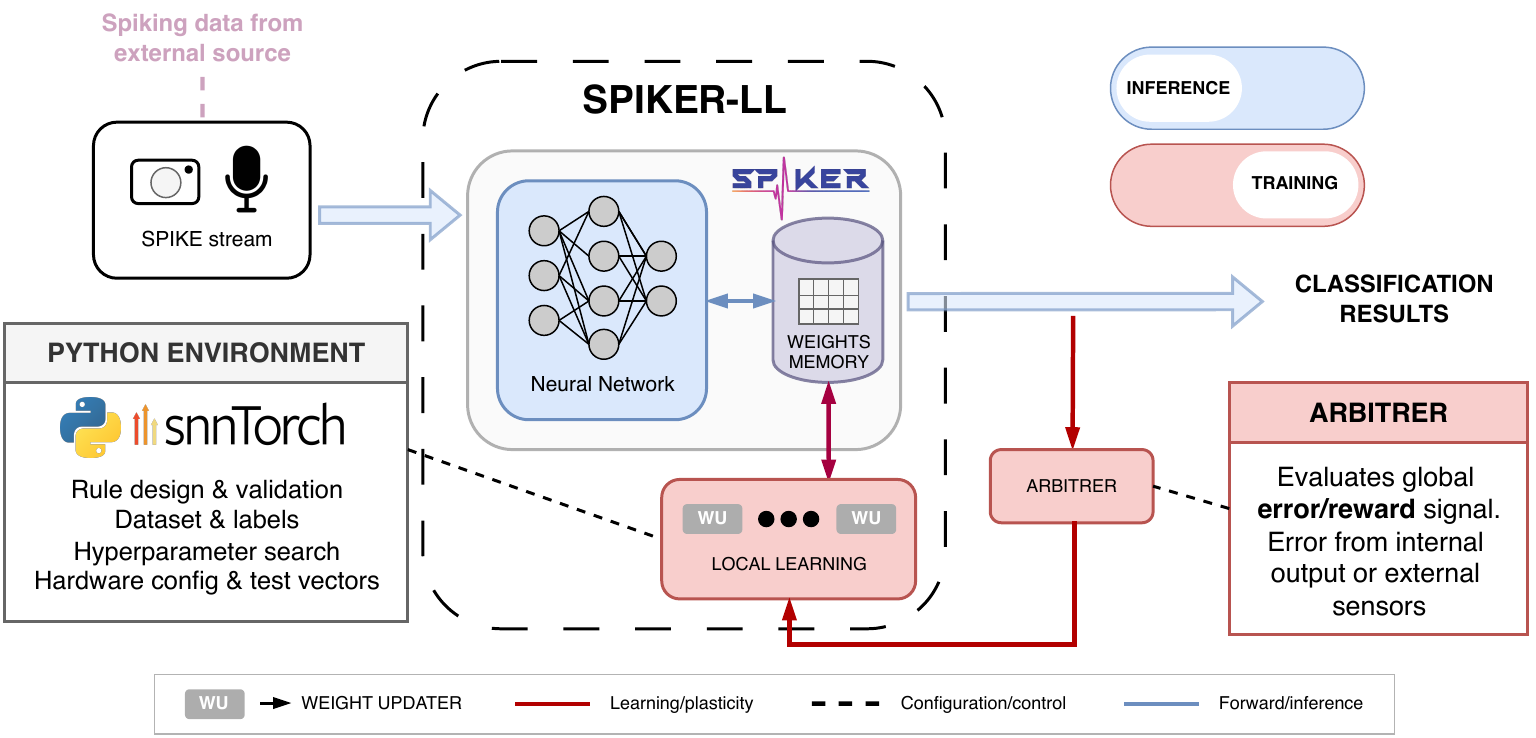}
  \caption{Envisioned deployment of \tool{} in the field with different operating modes.}
  \label{fig:vision}
\end{figure}

\subsection{\tool{} deployment environment}

Incorporating local learning into a hardware accelerator is not sufficient without a deployment workflow, illustrated in Figure~\ref{fig:vision}. \tool{} is instantiated by selecting the architectural configuration (e.g., number of layers, neurons, synapse density) through the Spiker+ VHDL generator, and processes spike-encoded inputs from neuromorphic front-ends (e.g., event-based cameras or digital cochleae).
\tool{} operates in two modes. Inference mode follows the standard high-throughput, low-latency Spiker+ pipeline. Training mode activates the local learning modules and is typically used during calibration or adaptation phases, exploiting the arbiter described in \autoref{sec:arbiter}. Because configuring learning parameters can be complex, \tool{} provides an snnTorch \gls{stsf} implementation that models hardware non-idealities (e.g., quantization, saturation, temporal gating). This framework enables rapid design-space exploration and hyperparameter tuning, and was validated by comparing signal traces between software and RTL simulations. The accelerator exposes an extended configuration interface allowing learning parameters to be updated at runtime.\footnote{The complete experimental code and implementation details are available at: \url{https://github.com/smilies-polito/Spiker}}

\section{Results}

The performance of \tool{} is analyzed along three axes: (i) behavior on representative benchmarks, (ii) overhead introduced by on-chip training, and (iii) comparison with prior work.

\begin{table*}[!tb]
  \centering
  \caption{Comparison between the proposed accelerator (Spiker-LL) and related SNN accelerators.}
  \label{tab:competitors}
  \footnotesize
  \setlength{\tabcolsep}{3pt}
  \resizebox{\textwidth}{!}{%
  \begin{tabular}{l *{9}{c}}
    \toprule
    & \textbf{Spiker-LL} & \textbf{Spiker-LL (8-bits)} & \textbf{\cite{11208669}} & \textbf{\cite{10841405}} & \textbf{\cite{10492616}} & \textbf{\cite{9822407}} & \textbf{\cite{leeSpikeTrainLevelDirect2020}} & \textbf{\cite{thaon.FPGABasedCoProcessorSpiking2022}} & \textbf{\cite{siddiqueLowCostNeuromorphic2023}} \\
    \midrule
    Year & \textbf{This work} & \textbf{This work} & 2025 & 2025 & 2024 & 2022 & 2020 & 2022 & 2023 \\
    Datasets & \textbf{MNIST} & \textbf{MNIST} & MNIST(scaled) & MNIST & MNIST & MNIST & MNIST (14x14) & Caltech-101 & DIGITS \\
    Learning rule & \textbf{STSF} & \textbf{STSF} & Integer E-Prop & Optimized AL & PLR & Triple Brain & Custom DFA & STDP & HaSiST \\
    Neuron type & \textbf{LIF} & \textbf{LIF} & IF & IF & LIF & LIF & LIF & IF & LIF \\
    Weight bw [bit] & \textbf{16} & \textbf{8} & 12 & 8 or 16 & 8 & 16 & 17 & 24 & 8 \\
    Neuron bw [bit] & \textbf{16} & \textbf{8} & 16 & 13 or 21 & 8 & 24 & 9 & 24 & 8 \\
    Platform & \textbf{Pynq Z2} & \textbf{Pynq Z2} & Zynq-7010 & Zynq-7045 & Zynq-7045 & Zynq-7045 & Xilinx ZC706 & Zynq-7000 Zedboard & Virtex-6 \\
    Architecture & \textbf{784-200-10} & \textbf{784-200-10} & 256-256-10 & 784-512-100 & 784-256 & 784-256 & 196-50-10 & DSNN w 3 conv/pool & 64-20-10 \\
    Accuracy [\%] & \textbf{92} & \textbf{90} & 97.55 & 97.21 & 90.19 & 95.10 & 94.34 & 95.7 & 99 \\
    Frequency [MHz] & \textbf{90} & \textbf{90} & 100 & 100 & 250 & 250 & 100 & 100 & 50 \\
    Power [mW] ([uW/syn]) & \textbf{509 (3.2)} & \textbf{326 (2.1)} & 139 (2.0) & 398 (0.88) & 637 (3.2) & 938 (4.7) & 113 (11) & 2,179 & -- \\
    \rowcolor{gray!20}
    Latency [ms] & \textbf{0.12} & \textbf{0.12} & 0.25 & 0.467 & -- & 0.74 & 3.998 & -- & -- \\
    \rowcolor{gray!20}
    Energy [mJ] ([nJ/syn]) & \textbf{0.061 (0.38)} & \textbf{0.039 (0.25)} & 0.05 (0.73) & 0.39 (0.86) & -- & 0.70 (3.5) & 0.452 (44) & -- & -- \\
    LUTs ([/syn]) & \textbf{13,312 (0.084)} & \textbf{8,459 (0.053)} & 8,755 (0.129) & 17,578 (0.039) & 6,033 (0.030) & 10,052 (0.050) & 33,484 (3.3) & 23,375 & (38) \\
    FFs ([$\times 10^{-3}$/syn]) & \textbf{6,144 (39)} & \textbf{4,429 (28)} & 2,970 (44) & 3,877 (8.6) & 1,747 (8.7) & 8,505 (42) & 6,836 (663) & 29,526 & (2.63$\times 10^{3}$) \\
    DSPs & \textbf{0} & \textbf{0} & 0 & 0 & 0 & 32 & 60 & 22 & -- \\
    BRAMs & \textbf{91.5} & \textbf{46} & 25.5 & 137.5 & 57 & 131 & -- & 97.5 & -- \\
    \bottomrule
  \end{tabular}
  }
\end{table*}

\subsection{Benchmarking}

\tool{} was evaluated on three standard SNN benchmarks (MNIST~\cite{6296535}, Fashion-MNIST (F-MNIST)~\cite{xiao2017fashionmnist}, and DIGITS~\cite{optical_recognition_of_handwritten_digits_80}) using Poisson rate coding over ten timesteps and 16-bit fixed-point weights and membrane potentials (eight fractional bits). Although hardware efficiency is the primary focus, hyperparameters were tuned per dataset, with a temporal gating of five timesteps consistently yielding the best accuracy. Table~\ref{tab:datasets} reports the resulting design metrics.
These results highlight that \tool{} is a modular and reconfigurable spiking architecture rather than a fixed-function accelerator. It supports different layer sizes, memory footprints, and learning parameters, enabling systematic exploration of accuracy--latency--area trade-offs while preserving the same hardware building blocks. Hardware metrics were generated using a standard Xilinx Vivado flow; power values rely on vector-less static analysis and are used as a best-effort estimate of intrinsic hardware cost for cross-paper FPGA comparisons.

The MNIST configuration represents a balanced mid-range design point, achieving sub-watt power, moderate resource use, and \(\sim\)8.3k images/s throughput at 90\,MHz, with low energy per inference suitable for edge deployment. Increasing the hidden layer from 200 to 300 neurons for F-MNIST demonstrates scalable model capacity: LUT/FF/BRAM usage grows predictably, while latency remains 0.12--0.13\,ms and frequency drops only slightly to 83\,MHz, preserving sub-millisecond operation. At the opposite end of the design space, the DIGITS configuration (64 input, 60 hidden neurons) fits in \(\sim\)5k LUTs and less than 20 BRAMs, yet still reaches 93\% accuracy at 125\,MHz. Across all configurations, no DSP blocks are used, making \tool{} appealing for low-cost or DSP-limited edge FPGAs.
Finally, Table~\ref{tab:datasets} reports training time per epoch on \tool{} and on a high-performance CPU (AMD Ryzen 7 5700U), highlighting the speedups that dedicated hardware can provide at constrained power.

\begin{table}[!th]
  \centering
  \caption{Spiker-LL benchmarking.}
  \label{tab:datasets}
  \footnotesize
  \setlength{\tabcolsep}{3pt}
  \begin{tabularx}{\columnwidth}{@{}l *{4}{Y}@{}}
    \toprule
    \textbf{Metric} & \textbf{MNIST} & \textbf{F-MNIST} & \textbf{DIGITS} \\
    \midrule
    Architecture                        & 784-200-10 & 784-300-10 & 64-60-10 \\
    Learning rate                       & 0.026      & 0.021      & 0.074 \\
    Beta                                & 0.875      & 0.875      & 0.5 \\
    Threshold                           & 1          & 1          & 1 \\
    Accuracy [\%]                       & 92         & 81         & 93 \\
    Frequency [MHz]                     & 90         & 83         & 125 \\
    Latency / image [ms]                & 0.121      & 0.133      & 0.0077 \\
    Average power [mW]                  & 509        & 690        & 277 \\
    Energy / image [mJ]                 & 0.0616     & 0.0918     & 0.00213 \\
    BRAM used                           & 92         & 137        & 17 \\
    DSP used                            & 0          & 0          & 0 \\
    LUTs used                           & 13,312     & 19,713     & 5,059 \\
    FFs used                            & 6,144      & 8,278      & 1,691 \\
    Epoch time (Acc- CPU) [s]      & 7.26 - 22  & 7.98 - 23  & 0.01 - 0.282 \\
    \bottomrule
  \end{tabularx}
\end{table}

\begin{table}[ht]
  \centering
  \caption{On-device learning overhead.}
  \label{tab:spiker_comparison}
  \footnotesize
  \begin{tabular}{lccc}
    \toprule
    \textbf{Metric}                     & \textbf{\tool} & \textbf{Spiker inference only}  \\
    \midrule
    Frequency [MHz]                     & 90     & 90      \\
    Latency / sample [ms]               & 0.121  & 0.104   \\
    Average power [mW]                  & 509    & 481     \\
    RAMB36 used                         & 91     & 35      \\
    RAMB18 used                         & 1      & 0       \\
    DSP used                            & 0      & 0       \\
    LUTs used                           & 13,337 & 10,916  \\
    FFs used                            & 6,146  & 6,068   \\
    \bottomrule
  \end{tabular}
\end{table}

\subsection{Hardware overhead}

Table~\ref{tab:spiker_comparison} shows the impact of adding training capabilities to the original Spiker inference architecture for the MNIST benchmark. As intended, the overhead is limited. The main contributors are the replacement of static single-port ROMs with dual-port RAMs to enable weight updates (RAMB36 usage increases from 35 to 91) and the local learning modules, which add about 22\% more LUTs. Despite these additions, power increases by less than 6\%.
Latency remains close to the inference-only design because \tool{} still processes spikes sequentially and training reuses the same compute pipeline. In the worst case, enabling weight updates at every timestep can at most double the cycle count; in practice, temporally gated updates (every 5 timesteps) substantially reduce this overhead, and the latency values in Table~\ref{tab:spiker_comparison} correspond to this typical configuration.

\subsection{Comparisons to related works}

Direct comparison with prior work is challenging because no existing accelerator implements \gls{stsf}, many designs are not fully open-source, and reported metrics are collected under heterogeneous datasets, network sizes, quantization, and platforms. We therefore provide a best-effort cross-paper comparison with representative accelerators in Table~\ref{tab:competitors}, and report \tool{} in both 16-bit and 8-bit configurations to partially account for quantization differences.
Overall, Table~\ref{tab:competitors} shows that \tool{} occupies a competitive point in the design space for commodity edge FPGAs: it remains DSP-free, achieves sub-millisecond latency, and provides low energy per inference with favorable energy-per-synapse scaling. Some works report higher MNIST accuracy, but typically on larger networks and/or higher-end devices; others target different learning rules or application settings, making one-to-one comparisons less meaningful. Several designs also omit latency or power measurements, which limits efficiency comparisons.
Among the closest FPGA references, MorphBungee-Tiny~\cite{11208669} achieves lower absolute power and resource usage with a smaller network, but with roughly \(2\times\) higher latency; \tool{} remains competitive in total energy per inference and improves energy per synapse. The edge processor in~\cite{10841405} reaches higher MNIST accuracy, but on a larger network and higher-end FPGA, with higher energy per inference and a substantially larger resource footprint. Spike-Train~\cite{leeSpikeTrainLevelDirect2020}, which uses a DFA-based rule related in spirit to our approach, reports higher latency and energy despite comparable frequency. These comparisons indicate that \tool{} favors a practical balance of efficiency, flexibility, and on-chip supervised learning support on commodity FPGAs rather than maximizing accuracy alone.

\section{Conclusion}
This work introduced an open-source accelerator enabling supervised on-chip learning via a hardware-adapted \gls{stsf} rule. By using coincidence-based updates without traces, \gls{stsf} avoids the memory and dataflow bottlenecks of more expensive learning rules, making on-device learning practical on small \glspl{fpga}. Our fixed-point, single-sample adaptation preserves the core learning dynamics with only marginal accuracy loss.
Although evaluated on MNIST-like datasets and from-scratch supervised training, the architecture is designed for broader adaptive scenarios through incremental updates, temporal gating, and runtime feedback. Future work will extend the framework to richer models and online adaptation settings, and further analyze stability and convergence under quantization and gating.


\end{document}